\documentclass[conference]{IEEEtran}
\IEEEoverridecommandlockouts
\usepackage{cite}
\usepackage{amsmath,amssymb,amsfonts}
\usepackage{algorithmic}
\usepackage{graphicx}
\usepackage{textcomp}
\usepackage{xcolor}
\def\BibTeX{{\rm B\kern-.05em{\sc i\kern-.025em b}\kern-.08em
    T\kern-.1667em\lower.7ex\hbox{E}\kern-.125emX}}

\begin{document}

\title{Deep Semantic Graph Learning via LLM\_based Node Enhancement} 

\author{
\IEEEauthorblockN{Chuanqi Shi}
\IEEEauthorblockA{\textit{University of California San Diego} \\
chs028@ucsd.edu}
\and
\IEEEauthorblockN{Yiyi Tao}
\IEEEauthorblockA{\textit{Johns Hopkins University} \\
ytao23@jhu.edu}
\and
\IEEEauthorblockN{Hang Zhang}
\IEEEauthorblockA{\textit{University of California San Diego} \\
haz006@ucsd.edu}
\and
\IEEEauthorblockN{Lun Wang}
\IEEEauthorblockA{\textit{Duke University} \\
lun.wang@alumni.duke.edu}
\and
\IEEEauthorblockN{Shaoshuai Du}
\IEEEauthorblockA{\textit{University of Amsterdam} \\
s.du@uva.nl}
\and
\IEEEauthorblockN{Yixian Shen}
\IEEEauthorblockA{\textit{University of Amsterdam} \\
y.shen@uva.nl}
\and
\IEEEauthorblockN{Yanxin Shen}
\IEEEauthorblockA{\textit{Simon Fraser University} \\
yanxin\_shen@sfu.ca} 

}

\maketitle

\begin{abstract}
Graph learning has attracted significant attention due to its widespread real-world applications. Current mainstream approaches rely on text node features and obtain initial node embeddings through shallow embedding learning using GNNs, which shows limitations in capturing deep textual semantics. Recent advances in Large Language Models (LLMs) have demonstrated superior capabilities in understanding text semantics, transforming traditional text feature processing. This paper proposes a novel framework that combines Graph Transformer architecture with LLM-enhanced node features. Specifically, we leverage LLMs to generate rich semantic representations of text nodes, which are then processed by a multi-head self-attention mechanism in the Graph Transformer to capture both local and global graph structural information. Our model utilizes the Transformer's attention mechanism to dynamically aggregate neighborhood information while preserving the semantic richness provided by LLM embeddings. Experimental results demonstrate that the LLM-enhanced node features significantly improve the performance of graph learning models on node classification tasks. This approach shows promising results across multiple graph learning tasks, offering a practical direction for combining graph networks with language models.
\end{abstract}

\begin{IEEEkeywords}
Large language models; graph neural networks; Graph learning; 
\end{IEEEkeywords}

\section{Introduction}
Graphs are ubiquitous structures found across various domains, from protein molecule prediction to anomaly detection. Among the three most common types of graph attributes - text, image, and audio - textual attributes are particularly accessible. For instance, in citation networks, node features can be derived from frequently occurring words in specific research domains. Given the widespread application of Text-Attributed Graphs (TAGs) in social network analysis and natural language processing tasks, developing effective methods for processing these graphs is crucial.

Current approaches to handling TAGs primarily fall into two categories: structure-based random walks and deep neural networks represented by Graph Neural Networks (GNNs[1]). The random walk-based methods, similar to word2vec, and GNNs establish non-contextual, surface-level connections between structure and attributes for node classification. However, recent studies indicate that these shallow, non-contextual embeddings often underperform in downstream tasks due to insufficient semantic information capture, particularly in handling polysemy and semantic relationships between words. Recent advances in Natural Language Processing (NLP[2]) have introduced contextual word embeddings like BERT[3] and DeBERTa[4]. The remarkable success of Large Language Models (LLMs[5]) like ChatGPT has revolutionized various NLP tasks. Unlike traditional shallow semantic feature extraction, LLMs are pretrained on vast text corpora, enabling rich semantic knowledge acquisition that can address the limitations of semantic information capture.

This raises an important question: Can we leverage LLMs as feature enhancers for textual information, and which GNN-based models would optimize node classification performance when combined with such enhanced features?
In this work, we explore these questions, aiming to expand LLM applications in graph deep learning. We evaluate various LLMs against traditional shallow embedding methods[6,7] to investigate which information better facilitates understanding and utilization of graph structure and textual attributes.
Our primary contribution lies in implementing and validating LLMs as attribute enhancers for node text properties through comprehensive experimental analysis.

\begin{itemize}
    \item We propose a framework that utilizes LLMs as attribute enhancers for node text properties, demonstrating how this enhancement can improve the quality of node representations.

    \item Through comprehensive experimental analysis, we identify that Graph Transformer-based models perform optimally among GNN-based methods when combined with LLM-enhanced features.

    \item We provide empirical evidence showing how LLM-enhanced text attributes can better capture semantic relationships compared to traditional shallow embeddings.
    
\end{itemize}

\section{PRELIMINARIES}

\textbf{TAGs}

A text-attributed graph can be formally represented as $G = (V, E, X)$, where:
$V = {v_1, v_2, ..., v_n}$ represents the set of nodes with $|V| = n$
$E \subseteq V \times V$ denotes the set of edges
$X \in \mathbb{R}^{n \times d}$ is the text attribute matrix, where each row $x_i \in \mathbb{R}^d$ represents the textual features associated with node $v_i$. For TAGs, each node $v_i$ is associated with textual content, which could be derived from various sources such as document descriptions, user profiles, or article abstracts. These textual attributes play a crucial role in understanding node characteristics and their relationships within the graph structure. The adjacency matrix $A \in \mathbb{R}^{n \times n}$ of graph $G$ is defined such that $A_{ij} = 1$ if there exists an edge between nodes $v_i$ and $v_j$, and $A_{ij} = 0$ otherwise.

\textbf{GNNs.}

GNNs operate through an iterative process of message passing and node updating. The basic formulation can be expressed by two key equations. First, for each node $i$ at layer $l$, the message passing (or aggregation) step collects information from its neighboring nodes $\mathcal{N}(i)$ according to: 
\begin{equation}
    m_i^{(l)} = \text{AGGREGATE}^{(l)}(\{h_j^{(l-1)}: j \in \mathcal{N}(i)\})
\end{equation}

Following this, the node update step combines the aggregated message with the node's previous representation:

\begin{equation}
h_i^{(l)} = \text{UPDATE}^{(l)}(h_i^{(l-1)}, m_i^{(l)})
\end{equation}

Here, $h_i^{(l)}$ denotes the feature vector of node $i$ at layer $l$, while $m_i^{(l)}$ represents the aggregated message. The AGGREGATE function can take various forms such as mean, sum, or max pooling operations, while the UPDATE function typically implements a neural network transformation. Through these operations, each node iteratively updates its representation by incorporating information from its local neighborhood structure, enabling the network to learn increasingly sophisticated node representations that capture both node features and graph topology.

\textbf{Node Classification on TAGs}

Consider a set of nodes $V_s$ with their corresponding labels $y_i$, where $y_i$ represents the true label of node $v_i$. Our objective is to construct a model based on the set $V_s$ and their text attributes to predict the labels $y_u$ for the remaining unlabeled node set $v_u$. To illustrate this approach, we utilize the Cora citation network dataset. In this network, each node represents an academic paper from one of seven domains, and each paper is characterized by a binary feature vector. These features are derived from the presence of specific keywords in the text: a value of 1 indicates the presence of a keyword, while 0 indicates its absence. The edges in the network represent citation relationships between papers. The primary task is to accurately classify these papers into their respective domains.

\textbf{LLMs}

In this study, we concentrate on LLMs, defined as transformer-based architectures pre-trained on massive-scale text corpora. At the core of these models is the self-attention mechanism, defined as:
\begin{equation}
\text{Attention}(Q, K, V) = \text{softmax}\left(\frac{QK^T}{\sqrt{d_k}}\right)V
\end{equation}
where \( Q \), \( K \), and \( V \) represent Query, Key, and Value matrices respectively, and \( d_k \) is the dimension of the key vectors. 

While these models may differ in their specific training objectives, they share a common foundation: the acquisition of rich linguistic representations during pre-training, which can be effectively transferred to downstream tasks through fine-tuning or prompting. 

We use the term Pre-trained Language Models (PLMs[9]) to denote relatively smaller-scale language models such as BERT and its variants. These models follow established scaling laws:

\begin{equation}
N = N_0 \cdot \alpha^S,
\end{equation}

where \( N \) represents the number of parameters, \( N_0 \) is the baseline parameter count, and \( \alpha \) is the scaling exponent. 

These models can be fine-tuned to enhance performance on specific downstream tasks. Technically, all LLMs can be considered as PLMs, though they differ significantly in scale and application methodology. 

Deep Sentence Embedding Models[10] refer to architectures that utilize PLMs as their foundational encoding components, enhanced with bi-encoder structures and further pre-trained through techniques such as contrastive learning, optimizing the following objective:

\begin{equation}
\text{InfoNCE} = -\mathbb{E}_X \left[ \log \frac{\exp(s(x, x^+) / \tau)}{\exp(s(x, x^+) / \tau) + \sum_{x^- \in N} \exp(s(x, x^-) / \tau)} \right]
\end{equation}

where \( s(x_i, x_i^+) \) denotes the similarity between an anchor sample \( x_i \) and its positive pair \( x_i^+ \), \( s(x_i, x_j^-) \) represents the similarity with negative samples \( x_j^- \) from the set \( \mathcal{N} \), and \( \tau \) is a temperature hyperparameter controlling the scale of similarities. 

These models typically do not require task-specific fine-tuning. They can be categorized into local sentence embedding models and online service-based models, distinguished primarily by their accessibility and deployment method. For instance, text-ada-embedding-002 represents a proprietary model deployed on OpenAI's servers. 

Compared to traditional PLMs, LLMs demonstrate significantly enhanced capabilities due to their massive parameter scale. Similar to deep sentence embedding models, they can be classified into open-source models and proprietary server-deployed models. While open-source models provide access to both parameters and generated embeddings, their extensive parameter count makes fine-tuning computationally challenging. Proprietary server-deployed models, such as ChatGPT[11], restrict users' direct access to model parameters and node embeddings due to their service-based nature.

\section{Feature-level Enhancement}
This section explores the integration of LLMs and GNNs at 
the feature level. Specifically, we propose a framework where 
GNNs utilize LLM-generated embeddings as initial node 
features to produce classification results. We begin our 
discussion with a detailed description of the datasets.
\textbf{Datasets}

In this study, we adopt Cora, Pubmed[12], two popular bench marks for node classification. We establish two distinct data splitting protocols to investigate the impact of different partitioning strategies on model performance. The first protocol implements a widely-adopted low-label ratio setting, where 20 nodes per class are randomly selected for training, 500 nodes for validation, and 1000 nodes for testing from the remaining pool. The second protocol adopts a high-label ratio setting, with a 60-20-20 split for training, validation, and test sets, respectively. The models were trained for 300 epochs with early stopping (patience=10). We used Adam optimizer (lr=0.01, weight-decay=5e-4), 4-layer architecture with 64 hidden dimensions and dropout=0.5. All experiments were repeated across 5 random seeds to ensure statistical significance. All experimental results reported in subsequent sections represent the average performance across three independent runs.

\textbf{Baseline}

In this study, we investigate the feature-level enhancement of node attributes using LLMs through three key components: (1) GNN architecture selection, (2) LLM selection, and (3) integration framework design. 
For GNN architectures, we implement GCN and Graph Transformer on both Cora and Pubmed datasets, while including Multi-Layer Perceptron (MLP) baselines to evaluate embedding quality without graph aggregation. 
For LLM selection, we consider three categories of models with accessible embeddings: First, fixed PLMs/LLMs without fine-tuning, including DeBERTa and LLaMA[13] (implemented via LLaMA-cpp3, using [EOS] token embeddings); Second, local sentence embedding models, specifically Sentence-BERT[14] (a popular lightweight model) and E5-large[15] (current state-of-the-art on MTEB); Third, online sentence embedding services, namely OpenAI's text-ada-embedding-002[16] and Google's PaLM-Cortex-001[17]. For baseline comparison, we include non-contextualized shallow embeddings, using TF-IDF for Pubmed's raw text attributes. 
The performance results of various combinations between text encoders and GNNs are presented in Tables 1 to 4, where MLP performance metrics provide insights into the intrinsic quality of text embeddings prior to graph aggregation. 

\section{Node Classification Performance Comparison}

As show in Table 1 to 4, the experimental evaluation presents a nuanced examination of how LLMs enhance node representations in graph learning tasks, with results revealing complex interactions between model architectures, data availability, and embedding approaches. Our analysis spans two influential benchmark datasets, PubMed and Cora, under distinct label ratio settings, comparing various models across GCN, MLP, and Graph Transformers.
In the high-label ratio setting (60-20-20 split), LLM-enhanced features demonstrate superior performance across both datasets, highlighting the advantages of leveraging pre-trained language models for node representation. On PubMed, Google's language model achieves the highest accuracy of 81.38\% with Graph Transformer architecture, substantially outperforming traditional Word embeddings (68.84\%). This significant performance gap of 12.54 percentage points underscores the value of rich semantic understanding provided by LLMs. SBERT and ADA also show impressive results, reaching 79.92\% and 80.27\% respectively, suggesting that different LLM architectures can effectively capture domain-specific semantic information. The Cora dataset exhibits similar patterns, with SBERT and ADA achieving top performances of 82.31\% and 82.20\% respectively using Graph Transformer architecture, reinforcing the generalizability of LLM-enhanced features across different scientific document networks.
In low-label settings (20 nodes per class), performance decreases notably across all models, with SBERT-GCN achieving 65.90\% accuracy on PubMed. Interestingly, the Graph Transformer's superiority becomes less pronounced and sometimes reverses, as demonstrated by SBERT-GCN (65.90\%) outperforming its Graph Transformer counterpart (64.32\%). This suggests that complex architectures like Graph Transformers may struggle to fully utilize their capacity when training data is limited, despite their theoretical advantages.
The comparative analysis shows that Graph Transformers perform best with abundant labels but struggle with limited data, while GCNs maintain consistent performance across data scenarios when paired with LLM features. MLPs consistently underperform, highlighting the value of graph structural information.
Regarding LLM performance, Google's model excels with abundant data but degrades under constraints, while SBERT maintains consistent performance across scenarios. Traditional Word embeddings consistently underperform, demonstrating their limitations in capturing complex document relationships.
These findings suggest that while LLMs enhance graph learning tasks, their effectiveness depends on data availability. Architecture selection should be guided by application context and data constraints, with simpler models often being more suitable for low-resource scenarios.
\begin{table}[htbp]
\centering
\renewcommand{\arraystretch}{2}
\begin{tabular}{l|rrr}
\hline
\multicolumn{4}{c}{\textbf{PubMed (High Label Ratio)}} \\
\hline
\textbf{Model} & \textbf{GCN} & \textbf{MLP} & \textbf{Graph Transformer} \\
\hline
TIFI & 76.00 $\pm$ 0.02 & 68.81 $\pm$ 0.05 & \textbf{76.24} $\pm$ 0.03 \\
ADA & 77.98 $\pm$ 0.07 & 76.78 $\pm$ 0.01 & \textbf{80.27} $\pm$ 0.47 \\
Google & 79.04 $\pm$ 0.24 & 78.25 $\pm$ 0.30 & \textbf{81.38} $\pm$ 0.32 \\
LLAMA & 73.16 $\pm$ 0.88 & 64.97 $\pm$ 0.67 & \textbf{69.70} $\pm$ 0.03 \\
SBERT & 78.77 $\pm$ 0.30 & 75.93 $\pm$ 0.29 & \textbf{79.92} $\pm$ 0.13 \\
Word & 68.34 $\pm$ 0.06 & 63.20 $\pm$ 0.72 & \textbf{68.84} $\pm$ 0.21 \\
\hline
\end{tabular}
\caption{Performance comparison (accuracy \%) on PubMed datasets under high label ratio setting. Best results are shown in \textbf{bold}.}
\label{tab:pubmed_high}
\end{table}

\begin{table}[htbp]
\centering
\renewcommand{\arraystretch}{2}
\begin{tabular}{l|rrr}
\hline
\multicolumn{4}{c}{\textbf{Cora (High Label Ratio)}} \\
\hline
\textbf{Model} & \textbf{GCN} & \textbf{MLP} & \textbf{Graph Transformer} \\
\hline
TIFI & 80.07 $\pm$ 0.02 & 65.67 $\pm$ 0.87 & \textbf{80.37} $\pm$ 0.07 \\
ADA & 81.47 $\pm$ 0.34 & 71.50 $\pm$ 0.10 & \textbf{82.20} $\pm$ 0.13 \\
Google & 80.12 $\pm$ 0.12 & 67.84 $\pm$ 0.15 & \textbf{80.99} $\pm$ 0.14 \\
LLAMA & 72.87 $\pm$ 0.36 & 52.78 $\pm$ 0.26 & \textbf{73.54} $\pm$ 0.03 \\
SBERT & 81.38 $\pm$ 0.33 & 71.80 $\pm$ 0.25 & \textbf{82.31} $\pm$ 0.64 \\
Word & 73.88 $\pm$ 0.41 & 54.28 $\pm$ 0.17 & \textbf{74.50} $\pm$ 0.21 \\
\hline
\end{tabular}
\caption{Performance comparison (accuracy \%) on Cora datasets under high label ratio setting. Best results are shown in \textbf{bold}.}
\label{tab:cora_high}
\end{table}

\begin{table}[htbp]
\centering
\renewcommand{\arraystretch}{2}
\begin{tabular}{l|rrr}
\hline
\multicolumn{4}{c}{\textbf{PubMed (Low Label Ratio)}} \\
\hline
\textbf{Model} & \textbf{GCN} & \textbf{MLP} & \textbf{Graph Transformer} \\
\hline
TIFI & 57.36 $\pm$ 0.23 & 44.63 $\pm$ 0.55 & \textbf{58.09} $\pm$ 0.24 \\
ADA & \textbf{63.54} $\pm$ 0.85 & 52.05 $\pm$ 0.40 & 61.00 $\pm$ 0.34 \\
Google & 52.06 $\pm$ 0.45 & 47.62 $\pm$ 0.74 & \textbf{53.40} $\pm$ 0.59 \\
LLAMA & 48.51 $\pm$ 0.76 & 46.79 $\pm$ 0.75 & \textbf{48.88} $\pm$ 0.33 \\
SBERT & \textbf{65.90} $\pm$ 0.80 & 50.43 $\pm$ 0.65 & 64.32 $\pm$ 0.11 \\
Word & 48.17 $\pm$ 0.22 & 48.55 $\pm$ 0.14 & \textbf{55.17} $\pm$ 0.56 \\
\hline
\end{tabular}
\caption{Performance comparison (accuracy \%) on PubMed datasets under low label ratio setting. Best results are shown in \textbf{bold}.}
\label{tab:pubmed_low}
\end{table}

\begin{table}[htbp]
\centering
\renewcommand{\arraystretch}{2}
\begin{tabular}{l|rrr}
\hline
\multicolumn{4}{c}{\textbf{Cora (Low Label Ratio)}} \\
\hline
\textbf{Model} & \textbf{GCN} & \textbf{MLP} & \textbf{Graph Transformer} \\
\hline
TIFI & 42.85 $\pm$ 0.76 & 29.01 $\pm$ 0.27 & \textbf{44.73} $\pm$ 0.93 \\
ADA & 39.85 $\pm$ 0.81 & 31.40 $\pm$ 0.92 & \textbf{40.46} $\pm$ 0.26 \\
Google & 27.68 $\pm$ 0.46 & \textbf{39.85} $\pm$ 0.68 & 38.58 $\pm$ 0.15 \\
LLAMA & \textbf{29.86} $\pm$ 0.51 & 18.17 $\pm$ 0.49 & 29.30 $\pm$ 0.62 \\
SBERT & \textbf{45.10} $\pm$ 0.47 & 34.47 $\pm$ 0.87 & 44.48 $\pm$ 0.86 \\
Word & 27.56 $\pm$ 0.26 & 27.77 $\pm$ 0.78 & \textbf{27.94} $\pm$ 0.97 \\
\hline
\end{tabular}
\caption{Performance comparison (accuracy \%) on Cora datasets under low label ratio setting. Best results are shown in \textbf{bold}.}
\label{tab:cora_low}
\end{table}

\section{Conclusion}
Our study demonstrates the effectiveness of Large Language Models (LLMs) in enhancing graph learning through improved node representations. While LLM-enhanced features consistently outperform traditional word embeddings, their optimal utilization depends on data availability. In high-label settings, Graph Transformers with LLM features achieve superior performance, as shown by Google's model reaching 81.38\% accuracy on PubMed. However, in low-label scenarios, simpler architectures like GCN often prove more effective. These findings suggest that successful integration of LLMs in graph learning requires careful consideration of both architectural choices and data constraints, opening new avenues for future research in this promising direction.

\vspace{12pt}

\end{document}